\documentclass[preprint,12pt]{elsarticle}
\usepackage{amsmath,graphicx,bm}
\usepackage{color}

\newcommand{\url}[1]{\textit{#1}}
\newcommand{\slot}[1]{\textsf{{\footnotesize #1}}}

\usepackage{amssymb}
\journal{Computer Speech and Language}

\begin{document}

\begin{frontmatter}



\title{Dialogue manager domain adaptation using Gaussian process reinforcement learning}


\author{Milica Ga{\v{s}}i\'c, Nikola Mrk{\v{s}}i\'c, Lina M. Rojas-Barahona, Pei-Hao Su, Stefan Ultes, David Vandyke, Tsung-Hsien Wen and  Steve Young}
\address{Trumpington Street, Cambridge CB2 1PZ\\
{\tt \{mg436,nm480,lmr46,phs26,su259,djv27,thw28,sjy\}@cam.ac.uk}}

\begin{abstract}
Spoken dialogue systems allow humans to interact with machines using natural speech. As such, they have many benefits. By using speech as the primary communication medium, a computer interface can facilitate swift, human-like acquisition of information. In recent years, speech interfaces have become ever more popular, as is evident from the rise of personal assistants such as Siri, Google Now, Cortana and Amazon Alexa. Recently, data-driven machine learning methods have been applied to dialogue modelling and the results achieved for limited-domain applications are comparable to or out-perform traditional approaches. Methods based on Gaussian processes are particularly effective as they enable good models to be estimated from limited training data.  Furthermore, they provide an explicit estimate of the uncertainty which is particularly useful for reinforcement learning. This article explores the additional steps that are necessary to extend these methods to model multiple dialogue domains. We show that Gaussian process reinforcement learning is an elegant framework that naturally supports a range of methods, including prior knowledge,  Bayesian committee machines and multi-agent learning, for facilitating extensible and adaptable dialogue systems.

\end{abstract}


\end{frontmatter}


\section{Introduction}
\label{sec:intro}
Spoken dialogue systems allow humans to interact with machines using natural speech. As such, they have many benefits. By using speech as the primary communication medium, a computer interface can facilitate swift, human-like acquisition of information. In recent years, systems with speech interfaces have become ever more popular, as is evident from the rise of personal assistants such as Siri, Google Now, Cortana and Amazon Alexa. Statistical approaches to dialogue management have been shown to reduce design costs and provide superior performance to hand-crafted systems particularly in noisy environments~\cite{ygtw13}. Traditionally, spoken dialogue systems were built for limited domains described by an underlying \emph{ontology}, which is essentially a structured representation of the database of entities that the dialogue system can talk about.

The semantic web is an effort to organise the large amount of information available on the Internet into a structure that can be more easily processed by a machine designed to perform reasoning on this data ~\cite{szlb14}. \emph{Knowledge graphs} are good examples of such structures. They typically consist of a set of triples, where each triple represents two entities connected by a specific relationship. Current knowledge graphs have millions of entities and billions of relations and are constantly growing. There has been a significant amount of work in spoken language understanding focused on exploiting knowledge graphs in order to improve coverage~\cite{tjwh12,hhtu13}. More recently there have also been efforts to build statistical dialogue systems that operate on large knowledge graphs, but limited so far to the problem of belief tracking~\cite{maps15,crms15}. In this article, we address the problem of decision-making in multi-domain dialogue systems. This a necessary step towards open-domain dialogue management. A previously proposed model for multi-domain dialogue management~\cite{whgt15} assumes a dialogue expert for each domain and the central controller which decides to which dialogue expert to pass the control. The dialogue experts are rule-based and the central controller is optimised using reinforcement learning. A related work in~\cite{wwss15} proposes a domain independent feature representation of the dialogue state so that the dialogue policy can be applied to different domains. Here, we explore multi-domain dialogue management which retains a separate statistical model for each domain.

Moving from a limited domain dialogue system that operates on a relatively modest ontology size to an open domain dialogue system that can converse about anything in a very large knowledge graph is a non-trivial problem. An open domain dialogue system can be seen as a (large) set of limited domain dialogue systems. If each of them were trained separately then an operational system would require sufficient training data for each individual topic in the knowledge graph, which is simply not feasible. What is more likely is that there will be limited and varied data drawn from different domains.  Over time, this data set will grow but there will always be topics within the graph which are rarely visited.

The key to statistical modelling of multi-domain dialogue systems is therefore the efficient reuse of data. Gaussian processes are a powerful method for efficient function estimation from sparse data. A Gaussian process is a Bayesian method which specifies a prior distribution over the unknown function and then given some observations estimates the posterior~\cite{rawi05}. A Gaussian process prior consists of a {\it mean function} - which is what we expect the unknown function to look like before we have seen any data - and the \emph{kernel function} which specifies the prior knowledge of the correlation of function values for different parts of the input space. For every input point, the kernel specifies the expected variation of where the function value will lie and once given some data, the kernel therefore defines the correlations between known and unknown function values. In that way, the known function values influence the regions where we do not have any data points. Also, for every input point the Gaussian process defines a Gaussian distribution over possible function values with mean and variance. When used inside a reinforcement learning framework, the variance can be used to guide exploration, avoiding the need to explore parts of the space where the Gaussian process is very certain. All this leads to very data efficient learning~\cite{gayo14}.

In this article, we explore how a Gaussian process-based reinforcement learning framework can be augmented to support multi-domain dialogue modelling focussing on three inter-related approaches. The first makes use of the Gaussian process prior. The idea is that where there is little training data available for a specific domain,  a \emph{generic} model can be used that has been trained on all available data. Then, when sufficient in-domain data becomes available, the generic model can serve as a prior to build a \emph{specific} model for the given domain. This idea was first proposed in~\cite{gktp15}. 

The second approach is based on a Bayesian committee machine~\cite{tre00}. The idea is that every domain or sub-domain is represented as a committee member. If each committee member is a Bayesian model, e.g.\ a Gaussian process, then the committee too is a Bayesian model, with mean and variance estimate. If a committee member is trained using limited data its estimates will carry a high uncertainty so the committee will rely on other more confident committee members, until it has seen enough training data. This method was proposed in~\cite{gmsv15}. It is similar to  Products of Gaussians which have previously been applied to problems such as speech recognition~\cite{gaai06}.

Finally, we extend the committee model to a multi-agent setting where committee members are seen as agents that collaboratively learn. This over-arching framework subsumes the first two approaches and provides a practical approach to on-line learning of dialogue decision policies for very large scale systems.  It constitutes the primary contribution of this article.

The remainder of the paper is organised as follows. In Section~\ref{sec:gpsdm}, the use of Gaussian process-based reinforcement learning (GPRL) is briefly reviewed.  The key advantage of GPRL in this context is that in addition to being data efficient, it directly supports the use of an existing model as a prior, thereby facilitating incremental adaptation.   In Section~\ref{sec:genpol},  various strategies for building a generic policy are considered and evaluated.
We then review the Bayesian committee machine in Section~\ref{sec:bcm}. Following that, in Section~\ref{sec:mdds}, we present a multi-domain dialogue manager based on the committee model. In Section~\ref{sec:mal}, we describe how multi-agent learning can be applied to the policy committee model. Then, in Section~\ref{sec:exp}, we present the experimental results. Finally, in Section~\ref{sec:concl}, conclusions together with future research directions are presented.

\section{Gaussian process reinforcement learning}\label{sec:gpsdm}

The input to a statistical dialogue manager is typically an N-best list of scored hypotheses obtained from the spoken language understanding unit. Based on this input, at every dialogue turn, a distribution of possible dialogue states called the \emph{belief state}, $\bm{b} \in \mathcal{B}$, an element of \emph{belief space}, is estimated. The belief state must accurately represent everything that happened prior to that turn in the dialogue. The quality of a dialogue is defined by a {\it reward function} $r(\bm{b},a)$ and the role of a dialogue policy $\pi$ is to map the belief state $\bm{b}$ into a system action $a \in \mathcal{A}$, an element of \emph{action space}, at each turn so as to maximise the expected cumulative reward. 

The expected cumulative reward for a given belief state $\bm{b}$ and action $a$ is defined by the $Q$-function:
\begin{equation}\label{eq:qfun}
Q(\bm{b},a)=E_{\pi}\left(\sum_{\tau=t+1}^{T}\gamma^{\tau-t-1}r_{\tau}|b_t=\bm{b}, a_t= a\right)
\end{equation}
where $r_{\tau}$ is the immediate reward obtained at time ${\tau}$, $T$ is the dialogue length and $\gamma$ is a discount factor, $0<\gamma\leq1$.
Optimising the $Q$-function is then equivalent to optimising the policy $\pi$.

GP-Sarsa is an on-line reinforcement learning algorithm that models the $Q$-function as a Gaussian process~\cite{ensh05}:
\begin{equation}
Q(\bm{b},a) \sim \mathcal{GP} \left (m(\bm{b},a),k((\bm{b},a),(\bm{b},a))\right)
\end{equation}
where $m(\cdot,\cdot)$ is the prior mean function and the kernel $k(\cdot,\cdot)$ is factored into separate kernels over  belief  and action spaces $k_{\mathcal{B}}(\bm{b}, \bm{b}')k_{\mathcal{A}}(a,a')$. 

For a training sequence of belief state-action pairs 
$\bm{B}=[(\bm{b}^0,a^0), \ldots,(\bm{b}^t,a^t)]^{\mathsf{T}}$ 
and the corresponding observed immediate rewards $\bm{r}=[r^1, \ldots, r^{t}]^{\mathsf{T}}$, the posterior of the $Q$-function for any belief state-action pair $(\bm{b},a)$ is given by:
\begin{equation}\label{eq:posterior}
 Q(\bm{b},a)|\bm{r},\bm{B} \sim \mathcal{N}(\overline{Q}(\bm{b},a),cov((\bm{b},a),(\bm{b},a)))
 \end{equation}
 where the posterior mean and covariance take the form:
 \begin{equation}\label{eq:posterior2}
 \begin{array}{ll}
&\overline{Q}(\bm{b},a)=\bm{k}(\bm{b},a)^{\mathsf{T}}\bm{H}^{\mathsf{T}}(\bm{H}\bm{K}\bm{H}^{\mathsf{T}}+\sigma^2\bm{H}\bm{H}^{\mathsf{T}})^{-1}(\bm{r}-\bm{m}),\\[2mm]
&cov((\bm{b},a),(\bm{b},a))=k((\bm{b},a),(\bm{b},a))-\\
&~~~\bm{k}(\bm{b},a)^{\mathsf{T}}\bm{H}^{\mathsf{T}}(\bm{H}\bm{K}\bm{H}^{\mathsf{T}}+\sigma^2\bm{H}\bm{H}^{\mathsf{T}})^{-1}\bm{H}\bm{k}(\bm{b},a)\\
\end{array}
\end{equation}
where  
$\bm{m}=[m(\bm{b}^0,a^0), \ldots, m(\bm{b}^t,a^t)]^{\mathsf{T}}$,
$\bm{k}(\bm{b},a)=[k((\bm{b}^0,a^0),(\bm{b},a)),\ldots,k((\bm{b}^t,a^t),(\bm{b},a))]^{\mathsf{T}}$, 
$\bm{K}$ is the Gram matrix~\cite{rawi05}, 
$\bm{H}$ is a band matrix with diagonal $[1, -\gamma]$ and
$\sigma^2$ is an additive noise factor which controls how much variability in the $Q$-function estimate is expected during the learning process. Since the Gaussian process for the $Q$-function defines a Gaussian distribution for every belief state-action pair (\ref{eq:posterior}), when a new belief point $\mathbf{b}$ is encountered, for each action $a \in \mathcal{A}$, there is a Gaussian distribution over $Q$-values.  Sampling from these Gaussian distributions gives $Q$-values 
%for each action $\{\hat{Q}(\mathbf{b},a) : a \in \mathcal{A} \}$ 
$\hat{Q}(\mathbf{b},a) \sim \mathcal{N}(\overline{Q}(\mathbf{b},a),\Sigma^Q(\bm{b},a))$ where $\Sigma^Q(\bm{b},a) = cov((\bm{b},a),(\bm{b},a))$ from which the action with the highest sampled $Q$-value can be selected:
\begin{equation}\label{eq:stoc}
\pi(\mathbf{b}) =\arg \max_{a} \left\{\hat{Q}(\mathbf{b},a) : a \in \mathcal{A} \right\}.
\end{equation}

% The likelihood of the sampled $\hat{Q}$ value is given by:
% \begin{equation}
% \label{eq:lik}
% \mathcal{L}(\hat{Q})=\frac{1}{\sqrt{2\pi\Sigma^Q(\bm{b},a))}}\exp\left(\frac{-|\overline{Q}(\bm{b},a)-\hat{Q}|^2}{2\Sigma^Q(\bm{b},a)}\right).
% \end{equation}

To use GPRL for dialogue, a kernel function must be defined on both the belief state space $\mathcal{B}$ and the action space $\mathcal{A}$. Here we use the Bayesian Update of Dialogue State (BUDS) dialogue model~\cite{thyo10}. The action space consists of a set of  slot-dependent and slot-independent summary actions. Slot-dependent summary actions include requesting the slot value, confirming the most likely slot value and selecting between top two slot values. Summary actions are mapped to  master actions using a set of rules and the kernel is defined as:
     \begin{equation}\label{eq:delta}
    k_{\mathcal{A}}(a, a')=\delta_a(a')
    \end{equation}
    
\noindent
where $\delta_a(a') = 1$ iff $a=a'$, $0$ otherwise.  The belief state consists of the probability distributions over the Bayesian network hidden nodes that relate to the dialogue history for each slot and the user goal for each slot. 
The dialogue history nodes can take a fixed number of values, whereas user goals range over the values defined for that particular slot in the ontology and can have very high cardinalities. 
User goal distributions are therefore sorted according to the probability assigned to each value since the choice of summary action does not depend on the values but rather on the overall shape of each distribution. 
The kernel function over both dialogue history and user goal nodes is based on the expected likelihood kernel~\cite{jekh04}, which is a simple linear inner product.
The kernel function for belief space is then the sum over all the individual hidden node kernels:
\begin{equation}\label{eq:fullker}
k_{\mathcal{B}}(\bm{b}, \bm{b}')=\sum_{h}\langle \bm{b}_h, \bm{b}_h' \rangle
\end{equation}
where $\bm{b}_h$ is the probability distribution encoded in the $h^{th}$ hidden node. 

\section{Distributed dialogue policies}\label{sec:genpol}
\begin{figure}[ht!]
\centering
\includegraphics[width=130mm]{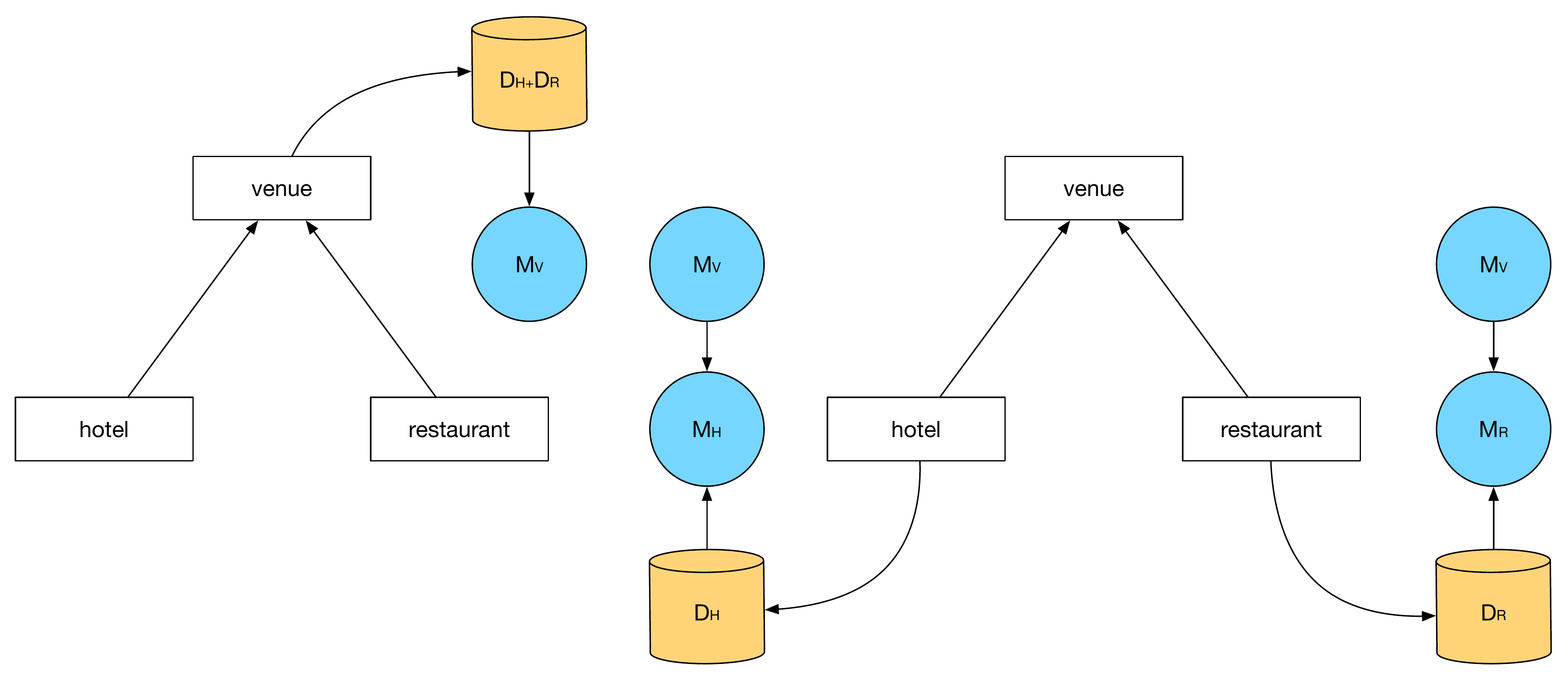}
\caption{Training a generic \slot{venue} policy model $M_{V}$ on data pooled from two subdomains $D_{R}+D_{H}$ (left);  and training specific policy models $M_{R}$  and $M_{H}$ using the generic policy $M_{V}$ as a prior and additional in-domain training data (right).}
\label{fig:dpd}
\end{figure}

One way to build a dialogue manager which can operate across a large knowledge graph is to decompose the dialogue policy into a set of topic specific policies that are distributed across the class nodes in the graph.   Initially, there will be relatively little training data and the system will need to rely on generic policies attached to high level generic class nodes which have been trained on whatever examples are available from across the pool of derived classes.   As more data is collected, specific policies can be trained for each derived class\footnote{cf  analogy with speech recognition adaptation using regression trees\cite{gale96}}.   An example of this is illustrated in Fig~\ref{fig:dpd}.  On the left side is the initial situation where conversations about hotels and restaurants are conducted using a generic model, $M_{V}$, trained on example dialogues from both the hotel and restaurant domains.   Once the system has been deployed and more training data has been collected, specific restaurant and hotel models $M_{R}$ and $M_{H}$ can be trained.\footnote{
Here a model $M$ is assumed to include input mappings for speech understanding, a dialogue policy $\pi$ and output mappings for generation.  
In this article, we are only concerned with dialogue management and hence the dialogue policy component $\pi$ of each model.
}

This type of multi-domain model assumes an agile deployment strategy which can be succinctly described as  ``deploy, collect data, and refine".  Its viability depends on the following assumptions:
\begin{enumerate}
\item it is possible to construct generic policies which provide acceptable user performance across a range of differing domains;
%, and better than can be obtained using under-trained domain specific policies;
\item as sufficient in-domain data becomes available, it is possible to seamlessly adapt the policy to improve performance, without subjecting users to unacceptable disruptions in performance during the adaptation period.
\end{enumerate}

In GPRL, the computation of $Q(\bm{b},a)$  requires the kernel function to be evaluated between $(\bm{b},a)$ and each of the belief-action points in the training data.  If the training data consists of dialogues from  subdomains (restaurants and hotels in this case) which have  domain-specific slots and actions, a strategy is needed for computing the kernel function between domains.  %Here three different strategies are considered.
%for building a generic policy from data from two subdomains. When a policy is trained on two subdomains the set of belief state-action pairs $\bm{B}$~(Eq.~\ref{eq:qfun}) consists of belief points and actions from both subdomains and the kernel function needs to deal with this situation.

If domains are organised in a class hierarchy it is expected that they share some of the slots. Calculating the kernel for shared parts of the belief state is straightforward:
%In this way, the kernel for belief state only consider hidden nodes that are mutual for both subdomains:
\begin{equation}\label{mutualker}
k_{\mathcal{B}}(\bm{b}^{\mathcal{H}}, \bm{b}^{\mathcal{R}})=\sum_{h \in \mathcal{H}\cap \mathcal{R} } \langle \bm{b}_h^{\mathcal{H}}, \bm{b}_h^{\mathcal{R}} \rangle,
\end{equation}
\noindent
where $\mathcal{R}$ and $\mathcal{H}$ are the considered subdomains.
When goal nodes are paired with differing cardinalities (eg \slot{name} might have different cardinality for different domains), the shorter vector is padded with zeros.  Pairing of
non-matching slots is achieved by treating them as {\it abstract slots}: \slot{slot-1}, \slot{slot-2}, etc so that they become the same in both subdomains according to some heuristics.
Hence for example, \slot{food} is matched with \slot{dogs allowed}, and so on.
%
%\begin{table}[h]
%\caption{Abstract slots for SFRestaurant and SFHotel subdomains}
%\begin{tabular}{ l |  l |l }
%\hline
%Abstract slot&SFRestaurant&SFHotel\\
%\hline
%slot1 & food & dogsallowed \\
%slot2 & goodformeal & hasinternet \\
%slot3 & kids-allowed &acceptscards\\
%\end{tabular}
%\end{table}
%In that way, the action space is the same for both subdomains and user goal nodes of the Bayesian network will also be the same so the kernel can be calculated as Eqs.~\ref{eq:fullker}~and~\ref{eq:delta}. 
As with the case with shared slots, when goal nodes are paired with differing cardinalities, the shorter vector is padded with zeros.
%can produce vectors of different length due to the difference in number of values an abstract slot can take, when the Eq.~\ref{eq:fullker} is applied 
%the shorter vector is a padded with zeros. 
%This strategy is referred to as \textbf{abstract}.
%Finally, a variation of the abstract strategy is considered where, instead of padding the shorter goal vectors with zeros,  the longer vectors are normalised to match the length of the shorter. We refer to this strategy as \textbf{renorm}. 
Other adaptation strategies are also possible 
but may result in increasing the dimensionality (see for example \cite{daume07}).

\section{Committee of dialogue policies}

\subsection{Bayesian committee machine}
\label{sec:bcm}
\begin{figure}[ht!]
\centering
\includegraphics[width=110mm]{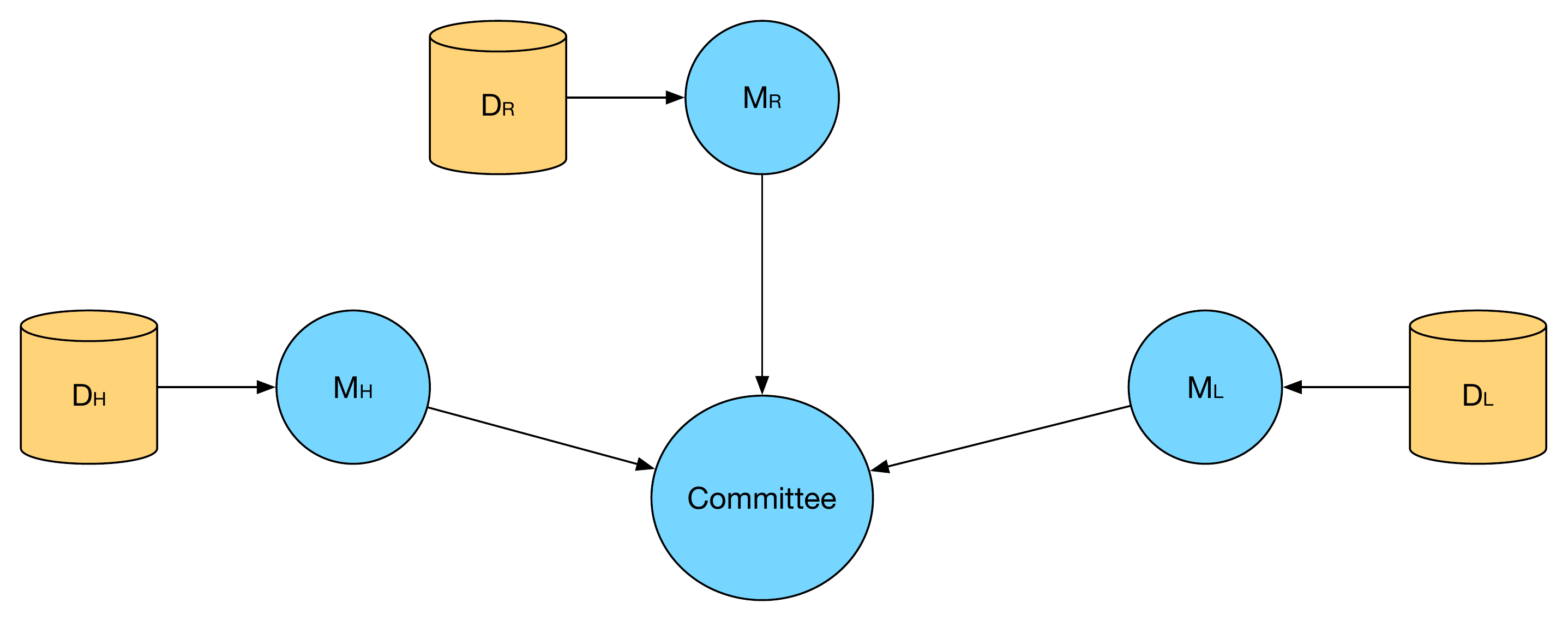}
\caption{Bayesian committee machine: Committee members consist of estimators trained on different datasets $D_i$. At every turn their estimated $Q$-values, $Q_i$, are combined to determine the final $Q$-value estimate.}
\label{fig:bcm}
%\vspace*{-2mm}
\end{figure}

The Bayesian committee machine is an approach to combining estimators that have been trained on different datasets. It is particularly suited to Gaussian process regression~\cite{tre00}. Here we apply the method to combine the outputs of multiple estimates of $Q$-values $Q_{i}$ with mean $\overline{Q}_i$ and covariance $\Sigma^Q_i$ as given by Eq.~\ref{eq:posterior2}.  Each estimator is trained on a distinct set of rewards and belief-state action pairs $\bm{r}_i,\bm{B}_i$ for $i \in \{1,\dots,M\}$, where $M$ is the number of policies in the policy committee.  As an example, Fig~\ref{fig:bcm} depicts a Bayesian committee machine consisting of three estimators.

Following the description in~\cite{tre00}, the combined mean $\overline{Q}$ and covariance $\Sigma^Q$ are calculated as:
 \begin{equation}\label{eq:bcm}
 \begin{array}{ll}
&\overline{Q}(\bm{b},a)=\Sigma^Q(\bm{b},a)\sum_{i=1}^{M}\Sigma^Q_{i}(\bm{b},a)^{-1}\overline{Q}_{i}(\bm{b},a),\\[2mm]
&\Sigma^Q(\bm{b},a)^{-1}=-(M\! -\! 1)*k((\bm{b},a),(\bm{b},a))^{-1} + \sum_{i=1}^{M}
\Sigma^Q_{i}(\bm{b},a)^{-1}.\\
\end{array}
\end{equation}

\vspace{8mm}

\subsection{Multi-domain Dialogue Manager}
\label{sec:mdds}

Section~\ref{sec:genpol}  introduced the notion of a \emph{generic} policy, which can be trained from data coming from different domains, and a \emph{specific} policy that can be derived from a generic policy using additional in-domain data. In order to produce a generic policy that works across multiple domains, a kernel function must be defined on belief states and actions that come from different domains. In that case, domains are organised in a class hierarchy so it is reasonable to assume that there are shared portions of the belief for different domains. These portions relate to shared slots and are directly mapped to each other and for slots which are different, the mapping can be defined either manually or by using some similarity metric.

\begin{figure}[ht!]
\centering
\includegraphics[width=140mm]{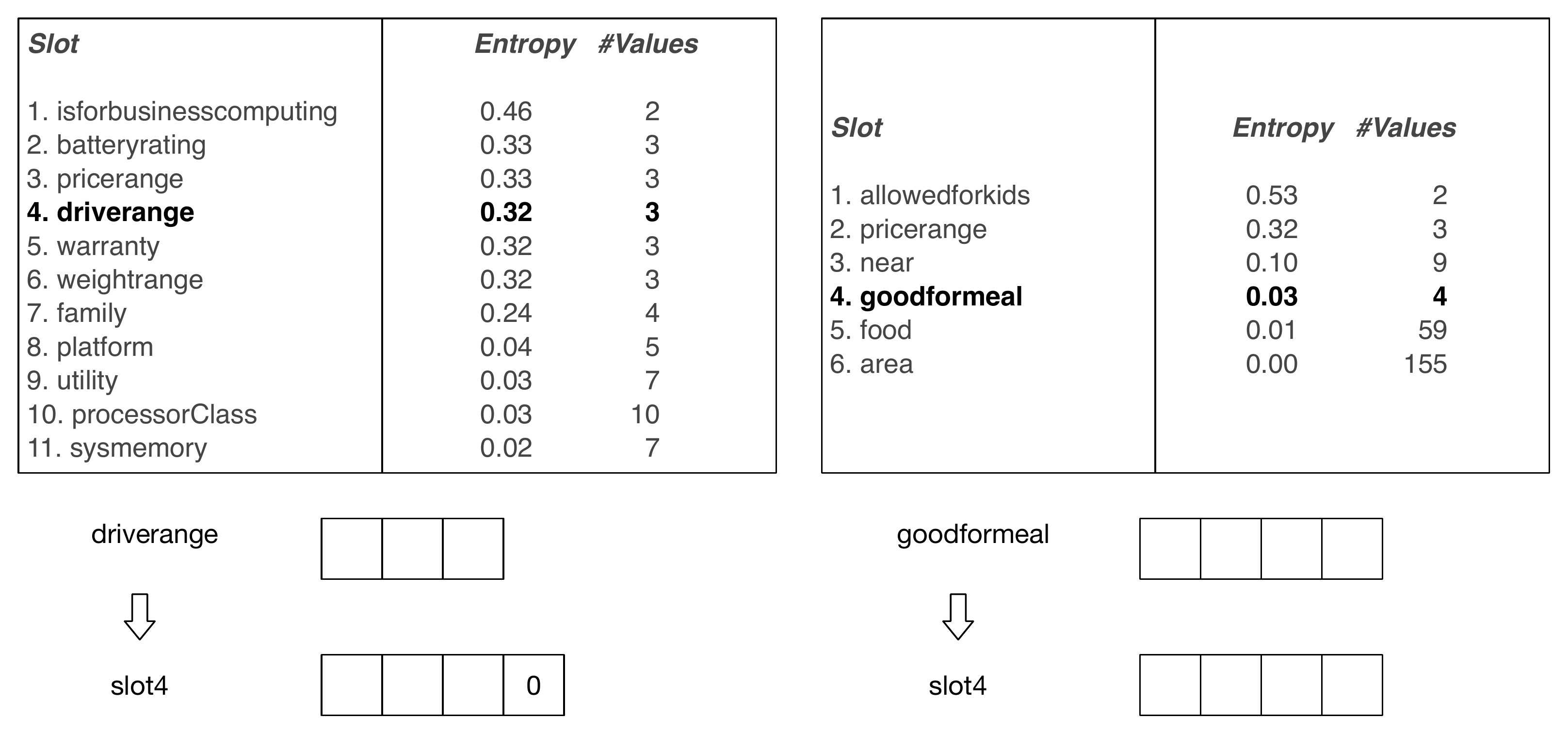}
\caption{Slot matching for slots that come from different domains for requestable slots semantic class. In this cases requestable slots for both domains are sorted according to their normalised entropy and then \slot{driverange} is being matched with \slot{goodformeal}. Since they have different number of values the shorter vector is padded with zeros.}
\label{fig:entropy}
%\vspace*{-2mm}
\end{figure}

When using a Bayesian committee machine, it is possible to have two domains which have no shared slots. Therefore, a different approach is required for building policies that can operate (and be trained on) belief states and actions that come from different domains. The approach is as follows. The slots from each domain are divided into semantic classes\footnote{In our case this was performed manually. For larger ontologies this could be induced from the ontology or another automatic method would be needed which is out of scope of this article.}. We have three semantic classes: 
\begin{description}
\item[name slot] refers to the name of the entity in the database;
\item[requestable slots] are the ones the user can specify to constrain a search, for instance slot \slot{food} or slot \slot{batteryrating};
\item[informable slots] are the ones the user can request further information, such as slot \slot{phone} or slot \slot{dimension}.
\end{description}
Then, the following steps are taken:
\begin{enumerate}
\item For each semantic class and for each slot in that semantic class,  the \emph{normalised entropy} $\eta$ is calculated by 
\begin{equation}
\label{eq:nent}
\eta(s)=-\sum_{v \in \mathcal{V}_s}\frac{p(s=v)\log(p(s=v))}{|\mathcal{V}_s|},
\end{equation}
where $s$ is a slot that takes values $v$ from a set $\mathcal{V}_s$ and where $p(s=v)$ is the empirical probability that an entity in the database with slot $s$ takes value $v$ for that slot. For example, if all entities in the database for the restaurant domain have \slot{area=centre}, then that slot has a normalised entropy equal to $0$. The measure is normalised so that slots that take different numbers of values can be compared. This measure provides an indicator of how useful each slot is in the dialogue. For instance, in this case it is not useful for the system to ask the user about their preference for slot \slot{area} since the answer provides no information gain.

\item For each domain, and for each semantic class, the slots are sorted based on their normalised entropy and given abstract names $slot^{c}_1,slot^{c}_2,\dots$ so that $\eta(slot^{c}_i)\geq\eta(slot^{c}_j)$ for $i\leq j$ for semantic class $c$.

\item The kernel function between belief states and actions which come from different domains $\mathcal{M}$ and $\mathcal{N}$, is calculated in the following way:
\begin{itemize}
\item Iteratively, for each $slot^c_i$ where $i \le min\{|M_c|,|N_c|\}$, index of the ordered list, in semantic class $c$ where $|M_c|$ denotes the number of slots in semantic class $c$ in domain $M$: \\
match the corresponding elements of belief space and actions, padding with zeros as necessary. 
\item Otherwise disregard the elements of the belief state relating to unpaired slots $j$ and if one of the actions relates to $slot_j$, consider the action kernel to be $0$.  
\end{itemize}
This slot matching process is illustrated in Figure~\ref{fig:entropy}.

%\item  When calculating the kernel function between belief states and actions which come from different domains $\mathcal{M}$ and $\mathcal{N}$, for each $i$ try to match portions of belief states  and actions related to $slot^{\mathcal{M}}_i$ from domain $\mathcal{M}$ to $slot^{\mathcal{N}}_i$ in domain $\mathcal{N}$ in the same semantic class, if both domains have the same slot $slot_i$. Otherwise, disregard the portion of the belief state relating to $slot_i$ and if one of the actions relates to $slot_i$, consider the action kernel to be $0$.  
\end{enumerate}

This approach has three important properties:
\begin{enumerate}
\item Once semantic classes are defined, the further process does not require human intervention to define the relationship between slots that come different domains;
\item it provides a well-defined computable relationship between any two domains; and
\item the kernel function that is defined in step 3 is positive definite so the Gaussian process is well-defined.
\end{enumerate}

\section{Multi-agent learning in the policy committee framework}
\label{sec:mal}
In the standard reinforcement learning framework there is a single agent that is trying to solve a specific task in a given environment. However, for complex tasks it has been shown~\cite{raic06} that it is more effective to decompose the problem into sub-tasks and introduce a distinct agent to solve each sub-task. In this case, each agent needs to take into account only part of the state space and this can significantly speed up the learning process. Learning  in such multi-agent systems is typically performed in three steps~\cite{raic06}. First, each agent proposes an action. Second, a gating mechanism, which can be either handcrafted or optimised automatically, is deployed to select the actual system action.  Finally, the reward is distributed among the agents and they each re-estimate their policy.

\begin{figure}[ht!]
\centering
\includegraphics[width=120mm]{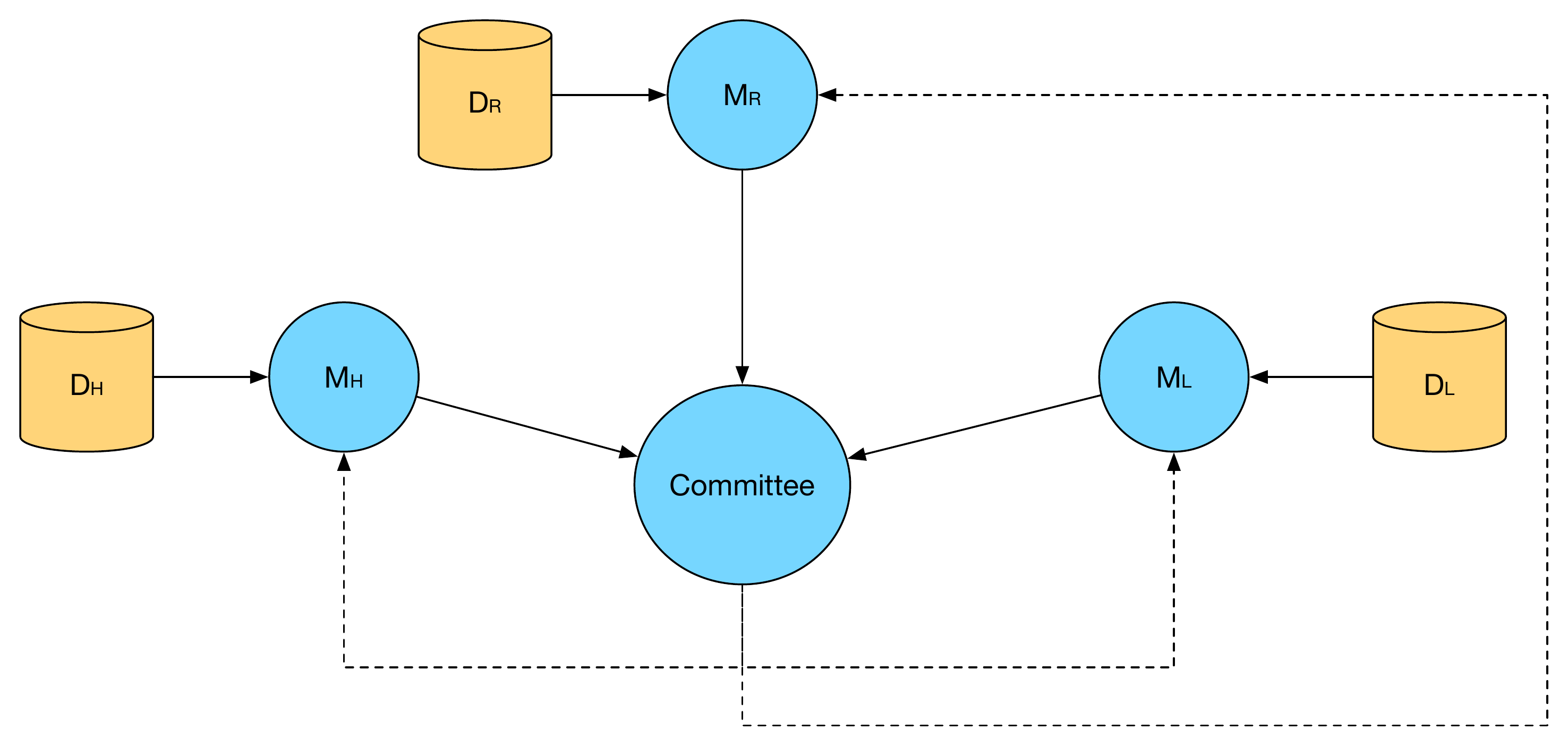}
\caption{Multi-agent policy committee model}
\label{fig:multiagent}
%\vspace*{-2mm}
\end{figure}

The multi-agent framework can be seen as an extension of the policy committee model (see Figure~\ref{fig:multiagent}). In fact, the first two steps are exactly the same: each committee member estimates its own $Q$-function and then Eq.~\ref{eq:bcm} is used as the gating to automatically combine the output. The multi-agent framework, however, includes a third step which is to distribute the reward so that each agent (i.e. committee member) can learn from every dialogue. 
%So far, the total reward was given only to the committee member that is estimating the $Q$-function for the current dialogue domain~\cite{gmsv15}. 
Intuitively, the reward should be given to the agent for the domain that the dialogue is currently in.  However, in practice it can be very difficult to identify a specific domain especially since the user can switch domains within the same dialogue. To avoid these issue, three strategies for distributing the reward are investigated:
\begin{description}
\item{na\"{i}ve approach:} The total reward that the system obtains is directly fed back to each committee member~\cite{raic06}.
\item{winner-takes-all approach:} The total reward that the system obtains is fed back to the committee member that proposed the highest $Q$-value for the action that was finally chosen by the gating mechanism~\cite{hump95}.
\item{reward scaling approach:} The total reward is redistributed to each committee member in such a way as to reflect its contribution to the final action chosen by the gating mechanism~\cite{raic06}.
\end{description}

\section{Experimental results}\label{sec:exp}

\subsection{Experimental set-up}\label{sec:setup}
In order to investigate the effectiveness of the methods discussed above, a variety of experimental contrasts were examined using an agenda-based simulated user operating at the dialogue act level~\cite{stwy07,kgjm10}.   The reward function allocates $-1$ at each turn  to encourage shorter dialogues, plus $20$ at the end of each successful dialogue.  The user simulator includes an error generator and this was set to generate incorrect user inputs $15$\% of time.

The proposed methods were also incorporated into a real-time spoken dialogue system in which policies were trained on-line using subjects recruited via Amazon Mturk. Each user was assigned specific tasks in a given subdomain and then asked to call the system in a similar set-up to that described in~\cite{jkgm11,gbhs13}. After each dialogue, the users were asked whether they judged the dialogue to be successful or not.  Based on that binary rating, the subjective success was calculated as well as the average reward. An objective rating was also computed by comparing the system outputs with the assigned task specification. During training, only dialogues where both objective and subjective score were the same were used.

% Dialogue policies were examined for three different domains. built for restaurants and hotels in San Francisco, USA, referred to as SFRestaurant and SFHotel respectively. SFRestaurant contains 239 entities and SFHotel has 182 entites. 
% These subdomains are described by ontologies that were automatically generated using information from the web~\cite{babg13}. A description of slots and the values that they take is given in Table~\ref{tab:domain}, where bold identifies the goal slots 
% that can be specified by the user and the remaining slots are informational slots that the user can query.

In order to examine the ability of the proposed methods to operate on multiple domains, four different domains were used: 
\begin{itemize}
\item[SFR] consisting of restaurants in San Francisco
\item[SFH] consisting of hotels in San Francisco
\item[L6] consisting of laptops with 6 properties that the user can specify
\item[L11] same as L6 but with 11 user-specifiable properties.
\end{itemize}
A description of each domain with slots sorted according to their normalised entropy is given in Table~\ref{tab:domain}.

\begin{table}[htbp]
\caption{Slots for each domain. The first section is slot \slot{name} which appears in each domain. The next section represents requestable slots, these are slots that can be specified by the user
to constrain a search. The remainder are informable slots which can be queried by the user regarding a specific entity.}
\begin{center}
\begin{tabular}{ |c| c| c | c| }
\hline
 SFR & SFH & L6 & L11  \\
\hline

\hline
 \slot{name}     &    \slot{name}  &\slot{name}   &      \slot{name}  \\
 \hline
  \slot{allowedforkids}  & \slot{dogsallowed}  & \slot{isforbusiness}  &  \slot{isforbusiness}   \\
 \slot{pricerange}  & \slot{pricerange}  & \slot{batteryratings}  &  \slot{batteryrating}   \\
 \slot{near}  & \slot{near} & \slot{pricerange}   &  \slot{pricerange}  \\
 \slot{goodformeal}  &  \slot{takescreditcards}  & \slot{driverange}             &  \slot{driverange}   \\
 \slot{food} &  \slot{hasinternet}    & \slot{weightrange}& \slot{weightrange} \\
 \slot{area} &  \slot{area}  &\slot{family}&\slot{family}  \\
 -                      &      - & -   & \slot{platform} \\
 -                      &      - &- & \slot{utility}\\
 -                       &     -  & -&\slot{processorclass} \\
 -                  &     -  &  -&\slot{sysmemory}\\
 \hline
 \slot{addr}   &   \slot{addr}&  \slot{price}&\slot{weight}\\
 \slot{price}   &    \slot{phone}&  \slot{drive}& \slot{battery}\\
 \slot{phone} &     \slot{postcode} & \slot{dimension}& \slot{price}\\
 \slot{postcode} &  -  &  -& \slot{dimension}\\
 - &  -  &  -& \slot{drive}\\
  - &  -  &  -& \slot{display}\\
   - &  -  &  -& \slot{graphadaptor}\\ 
    - &  -  &  -& \slot{design}\\
    - &  -  &  -& \slot{processor}\\ 
\hline
\end{tabular}
\label{tab:domain}
\end{center}
\end{table}

\subsection{Generic policy performance in simulation}\label{sec:genper}

In order to investigate the effectiveness of the generic policies discussed in Section~\ref{sec:genpol},  generic policies were trained and then tested in two domains -- SFR and SFH using equal numbers of restaurant and hotel dialogues.  In addition, in-domain policies were trained as a reference.
%we train a policy, interchangeably, on both subdomains, using an equal number of training dialogues for each subdomain. We then test each policy as a prior for each subdomain. In addition, we train two specific policies for each subdomain using the data only from that subdomain. 

%
%We first investigate the performance of these policies as priors for each subdomain~(Section~\ref{subsec:prior}) and then the way they influence the training of specific policies~(Section~\ref{subsec:tra}).
For each condition, 10 policies were trained using different random seeds and varying numbers of training dialogues. Each policy was then evaluated using $1000$ dialogues in each subdomain.   The overall average reward, success rate and  number of turns is given in Table~\ref{tab:prior} together with a 95\% confidence interval.  
%Bold values are statistically significant compared to non-bold values in the same group using an unpaired t-test with $p<0.01$. The only exception are the policies trained with $50000$ dialogues which are compared to the policies trained on $5000$ dialogues. 
The most important measure is the average reward, since the policies are trained to maximise this.

\begin{table}[htbp]
\begin{center}
\caption{Comparison of generic vs in-domain policies. Performance in each domain is measured in terms of reward, success rate and the average number of turns per dialogue. Results are given with a 95\% confidence interval. } 
\vspace{3mm}
\begin{tabular}{ l | r| r | r | r  } 
\hline
Strategy          & \#Dialogs                      &         Reward  &   Success  &   \#Turns \\
\hline
\multicolumn{5}{c}{SFRestaurant}  \\
\hline
in-domain &$250                               $& $   3.26 \pm 0.21$ & $  60.02 \pm 0.97$ & $   8.65 \pm 0.08$ \\
in-domain &$500                               $& $   5.00 \pm 0.21$ & $  68.17 \pm 0.91$ & $   8.55 \pm 0.07$ \\
%mutual    &$500                   $&$   3.88 \pm 0.21$ & $  62.60 \pm 0.95$ & $   8.56 \pm 0.07$ \\
generic   &$500                   $&$   4.48 \pm 0.21$ & $  67.35 \pm 0.92$ & $   8.89 \pm 0.08$ \\

%renorm &$500                 $&$   4.49 \pm 0.21$ & $  67.44 \pm 0.92$ & $   8.89 \pm 0.08$ \\

\hline
in-domain &$2500                      $&$   7.95 \pm 0.17$ & $  83.02 \pm 0.75$ & $   8.55 \pm 0.07$ \\
in-domain &$5000                      $&$   8.68 \pm 0.15$ & $  86.67 \pm 0.67$ & $   8.54 \pm 0.07$ \\
%mutual &$5000              $&$   8.28 \pm 0.16$ & $  84.63 \pm 0.71$ & $   8.52 \pm 0.07$ \\
generic &$5000             $&$   8.58 \pm 0.15$ & $  86.21 \pm 0.68$ & $   8.52 \pm 0.07$ \\
%renorm&$ 5000        $&$   8.60 \pm 0.16$ & $  86.32 \pm 0.68$ & $   8.56 \pm 0.07$ \\
\hline

\multicolumn{5}{c}{SFHotel}  \\
%\hline
%Name          &Tra diags                      &         Rew  &   Suc  &   Tur\\
\hline
in-domain &$250  $& $   3.58 \pm 0.21$ & $  62.07 \pm 0.96$ & $   8.75 \pm 0.07$ \\
in-domain &$ 500 $&$   4.83 \pm 0.21$ & $  69.08 \pm 0.92$ & $   8.89 \pm 0.08$ \\
%mutual&$ 500 $&$   4.81 \pm 0.21$ & $  68.00 \pm 0.92$ & $   8.71 \pm 0.08$ \\
generic&$ 500  $&$   5.27 \pm 0.20$ & $  70.01 \pm 0.90$ & $   8.64 \pm 0.07$ \\
%renorm &$500 $&$   4.64 \pm 0.21$ & $  68.59 \pm 0.92$ & $   8.96 \pm 0.08$ \\
\hline
in-domain &$2500         $&$   8.40 \pm 0.16$ & $  84.90 \pm 0.71$ & $   8.46 \pm 0.06$ \\
in-domain&$ 5000       $&$   8.92 \pm 0.15$ & $  87.48 \pm 0.65$ & $   8.45 \pm 0.06$ \\
%mutual &$5000   $& $   8.69 \pm 0.15$ & $  85.83 \pm 0.68$ & $   8.36 \pm 0.06$ \\
generic &$5000   $&$   8.89 \pm 0.15$ & $  87.19 \pm 0.66$ & $   8.44 \pm 0.06$ \\
%renorm &$5000 $&$   8.85 \pm 0.15$ & $  87.45 \pm 0.65$ & $   8.55 \pm 0.07$ \\
\hline

\end{tabular}
\label{tab:prior}
\end{center}
\end{table}

As can be seen from Table~\ref{tab:prior}, all generic policies perform better than the in-domain policies trained only on the data available for that subdomain (i.e. half of the training data available for the generic policy in this case) and this is especially the case when training data is limited. This suggests that the provision of generic policies in a large multi-domain will indeed provide robustness against the user moving into a domain for which there is very little training data.

%Secondly, in the majority of cases the strategies using abstract slots provide the best performance.  
%Thirdly, and somewhat surprisingly, the abstract generic policy yields better results than the in-domain policy even when the total amount of training data is the same. It may be that when limited data is available, varying subdomains increases exploration, leading to better performance. 
%Finally, training generic policies on more than $5000$ dialogues does not give a significant improvement in performance.

\subsection{Adaptation of in-domain policies using a generic policy as a prior in simulation}\label{sec:adapt}

We now investigate the effectiveness of using a generic policy as a prior for training an in-domain policy as in the right hand side of Fig.~\ref{fig:dpd}.  In order to examine the best and worst case,  the generic priors (from the 10 randomly seeded examples) that gave the best performance and the worst performance on each sub-domain trained with 500 and 5000 dialogues were selected. This results in four prior policies for each subdomain: generic-500-worst, generic-500-best, generic-5000-worst and generic-5000-best. 

\begin{figure}[ht!]
\centering
\includegraphics[trim = 0mm 5mm 0mm 5mm, width=130mm]{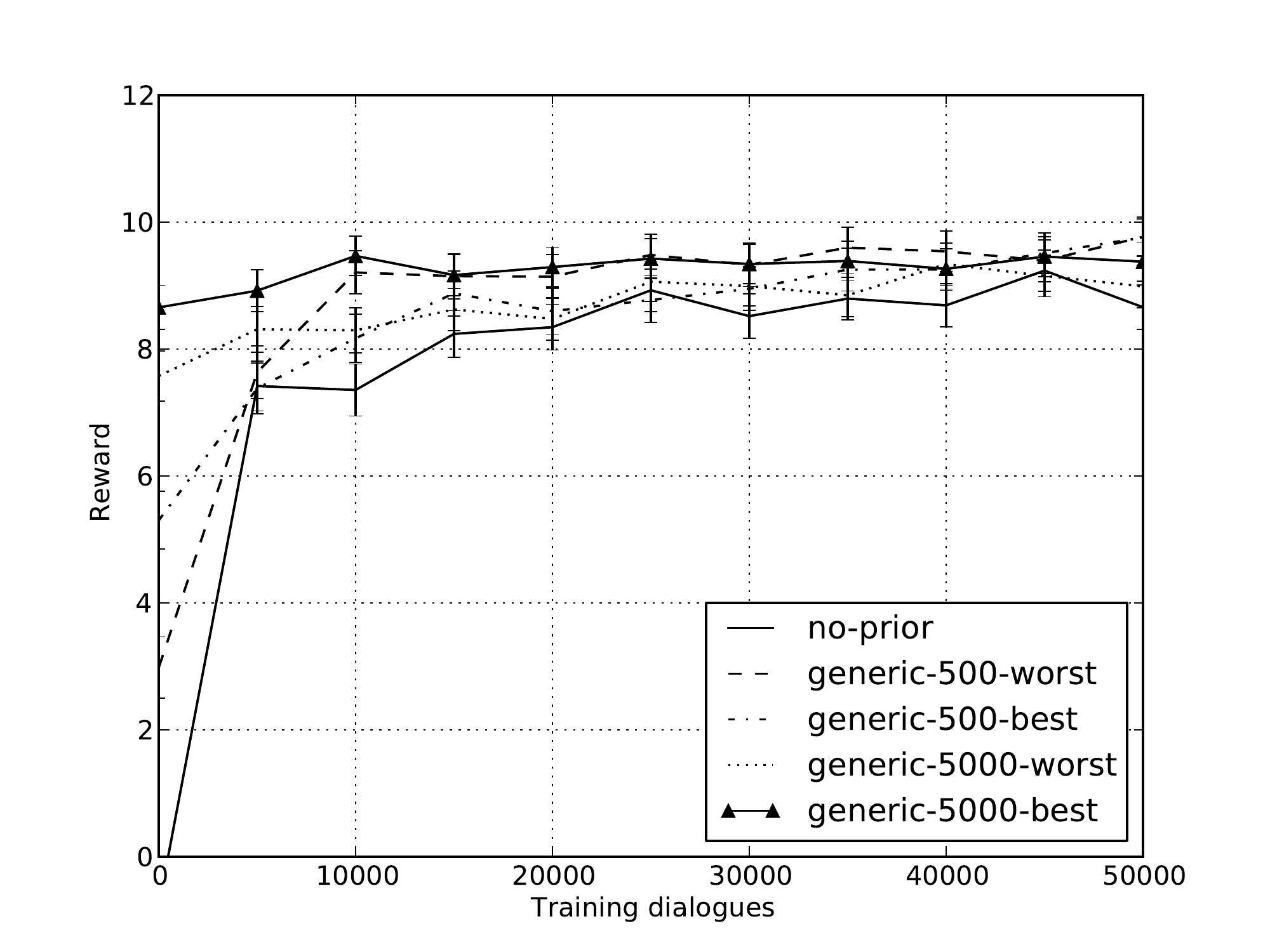}
\caption{Training policies with different priors -- SFR domain without prior and with different policies as prior. After every $1000$ dialogues the policies were evaluated with $1000$ dialogues and the average reward with a 95\% confidence interval is given on the figure. }
\label{fig:trainingSFRestaurant}
\end{figure}

\begin{figure}[ht!]
\centering
\includegraphics[trim = 0mm 5mm 0mm 5mm, width=130mm]{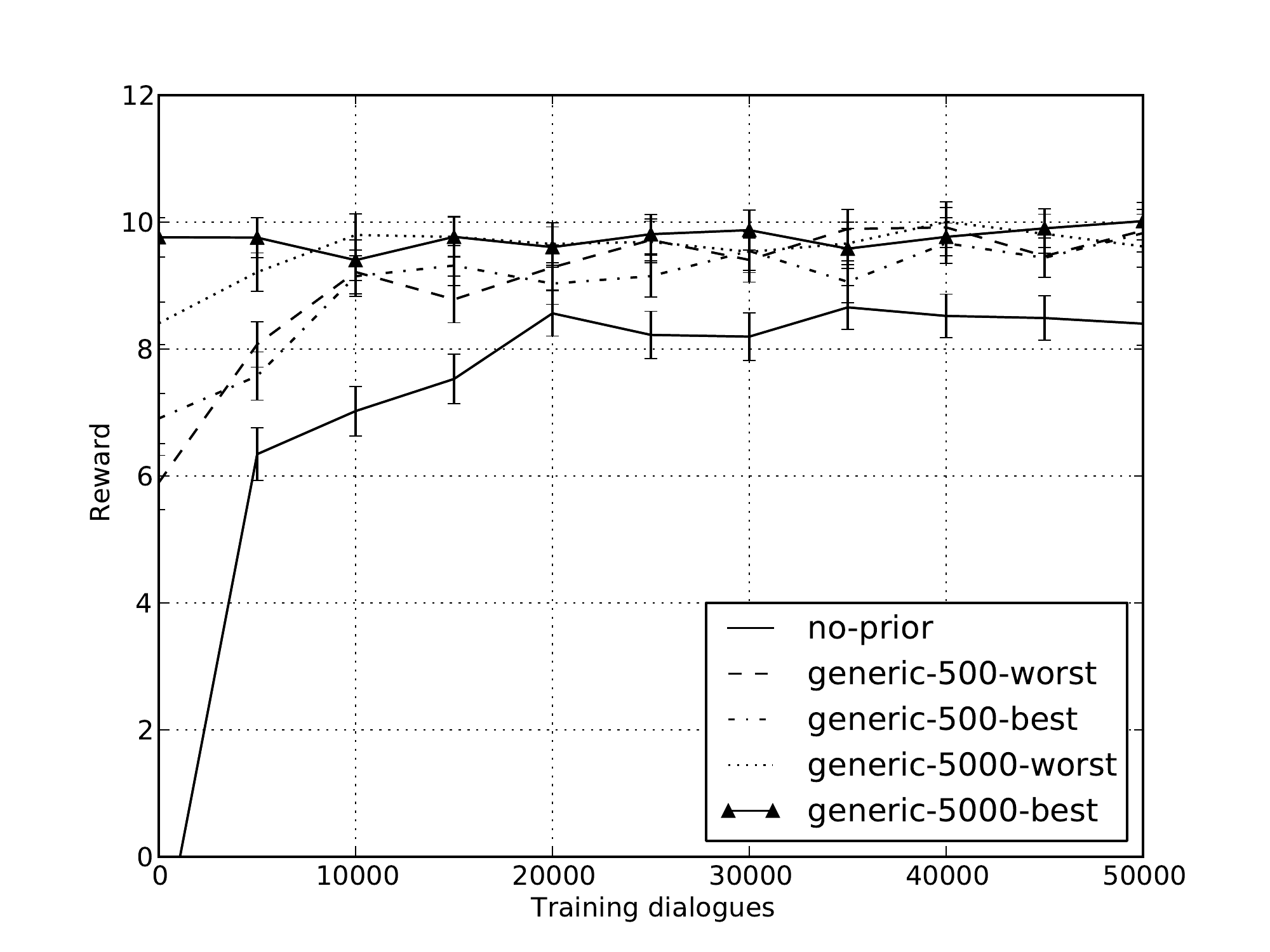}
\caption{Training policies with different priors -- SFH domain without prior and with different policies as prior. After every $1000$ dialogues the policies were evaluated with $1000$ dialogues and the average reward with a 95\% confidence interval is given on the figure.}
\label{fig:trainingSFHotel}
\end{figure}

In addition, a policy with no prior was also trained for each subdomain (i.e. the policy was trained from scratch). After every $1000$ training dialogues each policy was evaluated with $1000$ dialogues. The results are given in Fig.~\ref{fig:trainingSFRestaurant}~and~\ref{fig:trainingSFHotel} with a 95\% confidence interval.   Performance at $0$ training dialogues corresponds to using the generic policy as described in the previous section, or using a random policy for the no prior case.


\begin{table}[htbp]
\begin{center}
\caption{Performance  of best generic prior when adapted using 50K additional dialogues.  Results are given with 95\% confidence intervals.}
\vspace{3mm}
\begin{tabular}{ l | r | r | r  }
\hline
\multicolumn{4}{c}{SFR}  \\
\hline
Name                                &   Reward&    Success  &    \#Turns \\
\hline
best prior                   & $   8.66 \pm 0.35$ & $  85.40 \pm 2.19$ & $   8.32 \pm 0.20$ \\ 
adapted           & $   9.62 \pm 0.30$ & $  89.60 \pm 1.90$ & $   8.24 \pm 0.19$ \\
\hline
\multicolumn{4}{c}{SFH}  \\
\hline
best prior                 &$   9.76 \pm 0.31$ & $  88.80 \pm 1.96$ & $   7.95 \pm 0.21$ \\ 
adapted            & $  10.27 \pm 0.27$ & $  92.50 \pm 1.64$ & $   8.20 \pm 0.21$ \\
\end{tabular}
\label{tab:final}
\end{center}
\end{table}

The results from Figs.~\ref{fig:trainingSFRestaurant}~and~\ref{fig:trainingSFHotel} demonstrate that the performance of the policy in the initial stages of learning are significantly improved using the generic policy as a prior, even if that prior is trained on a small number of dialogues and even if it was the worst performing prior from the batch of 10 training sessions. These results also show that the use of a generic prior does not limit the optimality of the final policy. In fact, the use of a prior can be seen as resetting the variance of a GP which may lead to better sample efficiency~\cite{grwh14}.
This may be the reason why in some cases, the no-prior policies never catch up with the adapted policies - as in Figure~\ref{fig:trainingSFHotel}.

In Table~\ref{tab:final}, the performance of the best performing generic prior is compared to the performance of the same policy adapted using an additional 50K dialogues.  The results show that additional in-domain adaptation has the potential to improve the performance further.   So when enough training data is available, it is beneficial to create individual in-domain policies rather than continuing to train the generic policy. % This may be an example of the fact that  optimal performance can only be reached when the training and testing conditions match.

\subsection{Adaptation in interaction with human users}\label{sec:humres}

%They were asked to find restaurants that have particular features as defined by the given task. 

\begin{figure}[ht!]
\centering
\includegraphics[trim = 0mm 5mm 0mm 5mm, width=100mm]{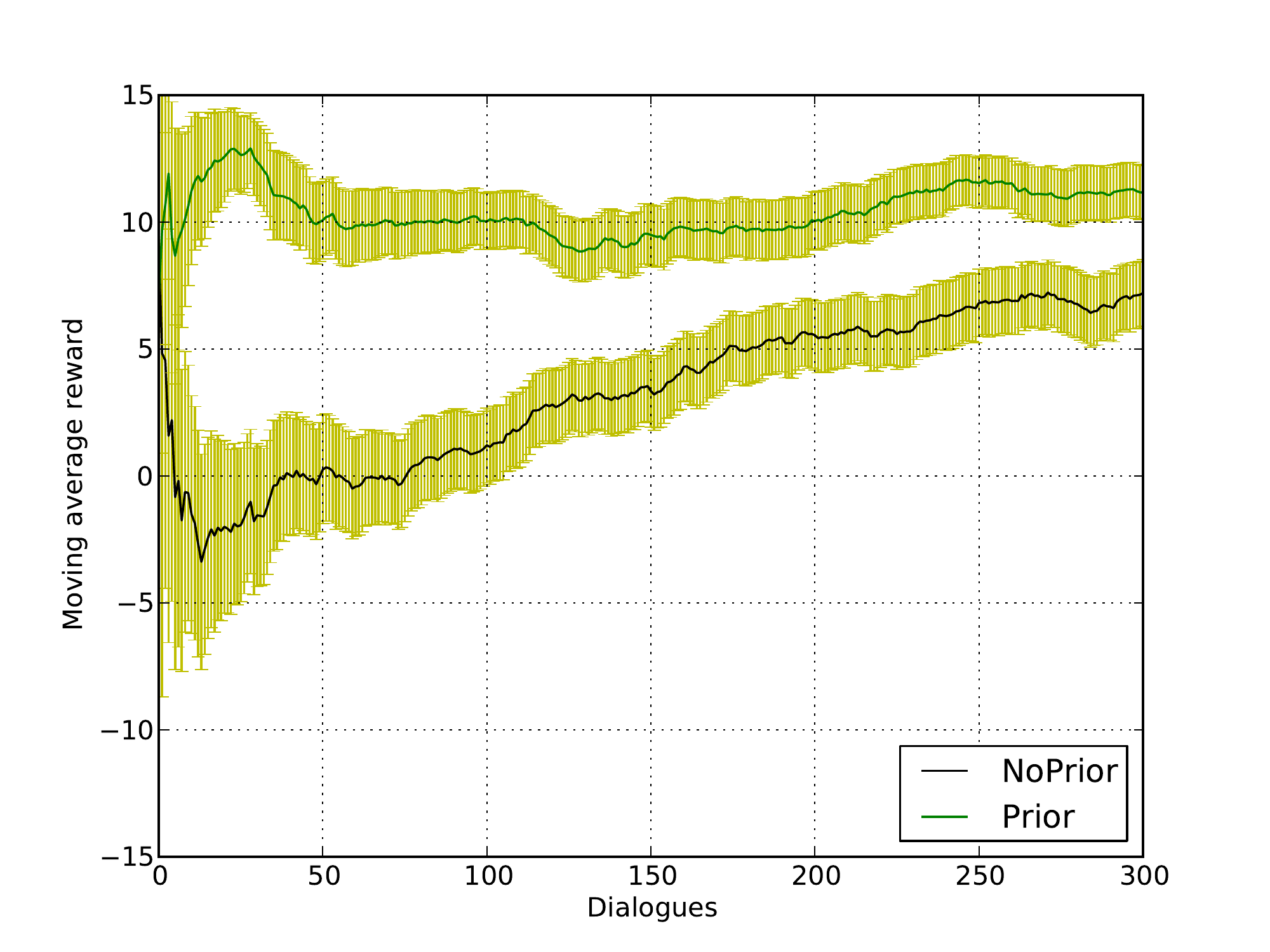}
\caption{Training in interaction with human users on SFR -- moving average reward. Plots are average of three sample runs.}
\label{fig:humtrarew}
\end{figure}

To examine performance when training with real users, rather than a simulator,
two training schedules were performed in the SFR subdomain -- one training from scratch without a prior, and the other performing adaptation using the best generic prior obtained after $5000$ simulated training dialogues. For each training schedule three sample runs were performed and the results were averaged to reduce any random variation. Fig.~\ref{fig:humtrarew} shows the moving average reward as a function of the number of training dialogues. The moving window was set to $100$ dialogues so that after the initial $100$ dialogues each point on the graph is an average of $300$ dialogues (3 sample runs $\times$ window size). The shaded area represents a 95\% confidence interval.   The initial parts of the graph exhibit more randomness in behaviour because the number of training dialogues is small.

The results show an upward trend in performance particularly for the policy that uses no prior.  However, the performance obtained with the prior is significantly better than without a prior both in terms of the reward and the success rate.   Equally importantly,  unlike the system trained from scratch with no prior,  the users of the adapted system are not subjected to poor performance during the early stages of training.

\subsection{Policy committee evaluation with simulated user}

In the previous section, the benefit of training generic models was demonstrated when training data is sparse. Here we investigate whether the use of a Bayesian committee machine can improve robustness further. The contrasts studied were as follows:

\begin{description}
\item{INDOM} \textbf{In-domain policy} -- trained only on in-domain data, other data is not taken into consideration, action-selection is based only on the in-domain policy. This is the baseline.
\item{GEN} \textbf{Single generic policy} -- one policy trained on all available data~(as in Section~\ref{sec:genpol}).
% \item{2CQ} \textbf{Two-policy committee with $Q$-values} -- one generic policy trained on all available data and one specific policy trained only on in-domain data.  Action selection depends directly on each $Q$-value i.e.\ the action that has the highest $Q$-value between the two policies is taken.
% \item{2CQL} \textbf{Two-policy committee with $Q$-values and likelihood} -- same as 2CQ but the Q values are scaled by the likelihood (Eq.~\ref{eq:lik}) for action selection.
% \item{2BCM} \textbf{Two-policy Bayesian committee machine} -- same as 2CQ but uses a Bayesian committee machine described in Section~\ref{sec:bcm} to provide the consensus estimate of the $Q$-value for action selection.
\item{MBCM} \textbf{Multi-policy Bayesian committee machine} -- as described in Section~\ref{sec:bcm}.  There is one committee member for each domain and each committee member is trained only on in-domain data.  However, for action-selection, the estimates of all committee members are taken into account using Eq~\ref{eq:bcm}, both during training and testing.
\item{GOLD} \textbf{Gold standard} -- this is the performance of the single policy where all training data comes from the same domain i.e.\ for $N$ domains, GOLD has $N$ times the number of in-domain dialogues for training as provided to INDOM.
\end{description}
 
We examine two cases: when the training data is limited, with only $250$ dialogues available for each domain, and when there is more training data available, $2500$ for each domain. In the evaluation of generic policies in Section~\ref{sec:genper}, the test domains were relatively similar. Here, we consider more diverse domains:
\begin{itemize}
\item Multi-domain system for SFR, SFH and L6, where the domains have different slots but each domain has the same number of slots, and
\item Multi-domain system for SFR, SFH and L11, where not only are the slots different, but also one of the domains, L11, has many more slots than the others.
\end{itemize}

For each contrast described above, 10 policies were trained on the simulated user using different random seeds. Each policy was then evaluated using $1000$ dialogues in each domain.   The overall average reward, success rate and  number of turns are given in Table~\ref{tab:res} together with 95\% confidence intervals. We do not report results on SFH domain as policies on this domain behave similarly to the ones on SFR domain (see Fig.~\ref{fig:trainingSFRestaurant}~and~\ref{fig:trainingSFHotel}).

\begin{table}[htbp]

\caption{Comparison of strategies for multi-domain adaptation. In-domain performance is measured in terms of reward, success rate and the average number of turns per dialogue. Results are given with 95\% confidence intervals. } 
\begin{center}
\begin{tabular}{ l | r | r | r  } 
\hline
Strategy                     &         Reward  &   Success  &   \#Turns \\
\hline
\hline
\multicolumn{4}{c}{L6 trained on $750$ dialogues from SFR, SFH, L6 }  \\
\hline
INDOM   & $   7.92 \pm 0.20$ & $  72.64 \pm 0.87$ & $   6.56 \pm 0.07$ \\
GEN  & $   9.34 \pm 0.19$ & $  79.43 \pm 0.80$ & $   6.49 \pm 0.06$ \\
% 2CQ   & $   8.95 \pm 0.20$ & $  78.91 \pm 0.80$ & $   6.77 \pm 0.07$ \\

% 2CQL & $   9.69 \pm 0.18$ & $  81.87 \pm 0.76$ & $   6.65 \pm 0.07$ \\

% 2BCM  & $  10.22 \pm 0.17$ & $  84.30 \pm 0.71$ & $   6.62 \pm 0.07$ \\
MBCM  & $   9.89 \pm 0.18$ & $  82.95 \pm 0.74$ & $   6.68 \pm 0.07$ \\
GOLD & $   9.25 \pm 0.19$ & $  80.35 \pm 0.79$ & $   6.77 \pm 0.07$ \\
\hline
\multicolumn{4}{c}{L6 trained on $7500$ dialogues from SFR, SFH, L6 }  \\
\hline
INDOM  & $  10.62 \pm 0.16$ & $  86.04 \pm 0.68$ & $   6.50 \pm 0.06$ \\
MBCM & $  11.60 \pm 0.14$ & $  90.32 \pm 0.58$ & $   6.42 \pm 0.06$ \\
GOLD  & $  11.98 \pm 0.13$ & $  92.36 \pm 0.53$ & $   6.42 \pm 0.06$ \\
\hline
\hline
\multicolumn{4}{c}{SFR trained on $750$ dialogues from SFR, SFH, L11 }  \\
\hline
INDOM  & $   5.73 \pm 0.21$ & $  68.17 \pm 0.92$ & $   7.89 \pm 0.08$ \\
GEN & $   6.32 \pm 0.21$ & $  72.04 \pm 0.89$ & $   8.05 \pm 0.08$ \\
% 2CQ   & $   5.73 \pm 0.21$ & $  70.13 \pm 0.90$ & $   8.24 \pm 0.08$ \\
% 2CQL & $   6.78 \pm 0.21$ & $  75.36 \pm 0.85$ & $   8.26 \pm 0.09$ \\

% 2BCM  & $   6.46 \pm 0.22$ & $  73.80 \pm 0.87$ & $   8.25 \pm 0.09$ \\
MBCM  & $   7.37 \pm 0.20$ & $  76.60 \pm 0.83$ & $   7.92 \pm 0.08$ \\
GOLD & $   7.34 \pm 0.20$ & $  76.97 \pm 0.83$ & $   8.01 \pm 0.08$ \\
\hline
\multicolumn{4}{c}{SFR trained on $7500$ dialogues from SFR, SFH, L11 }  \\
\hline
INDOM  & $   9.03 \pm 0.17$ & $  85.16 \pm 0.70$ & $   7.97 \pm 0.08$ \\
MBCM & $   9.67 \pm 0.17$ & $  88.28 \pm 0.66$ & $   7.96 \pm 0.08$ \\
GOLD  & $   9.65 \pm 0.16$ & $  88.80 \pm 0.62$ & $   8.05 \pm 0.08$ \\
\hline
\hline
\multicolumn{4}{c}{L11 trained on $750$ dialogues from SFR, SFH, L11 }  \\
\hline
INDOM  & $   6.46 \pm 0.22$ & $  67.59 \pm 0.92$ & $   7.02 \pm 0.08$ \\
GEN& $   7.18 \pm 0.21$ & $  70.91 \pm 0.89$ & $   6.97 \pm 0.08$ \\
% 2CQ  & $   6.91 \pm 0.21$ & $  70.15 \pm 0.90$ & $   7.09 \pm 0.08$ \\
% 2CQL & $   7.24 \pm 0.22$ & $  72.24 \pm 0.88$ & $   7.17 \pm 0.08$ \\

% 2BCM & $   6.55 \pm 0.23$ & $  69.11 \pm 0.91$ & $   7.20 \pm 0.09$ \\
MBCM  & $   8.52 \pm 0.20$ & $  77.09 \pm 0.82$ & $   6.88 \pm 0.07$ \\
GOLD & $   8.68 \pm 0.20$ & $  77.26 \pm 0.83$ & $   6.74 \pm 0.07$ \\
\hline
\multicolumn{4}{c}{L11 trained on $7500$ dialogues from SFR, SFH, L11 }  \\
\hline
INDOM & $  10.05 \pm 0.17$ & $  84.58 \pm 0.71$ & $   6.84 \pm 0.07$ \\
MBCM  & $  10.73 \pm 0.16$ & $  87.23 \pm 0.66$ & $   6.70 \pm 0.07$ \\
GOLD  & $  11.17 \pm 0.15$ & $  88.89 \pm 0.62$ & $   6.57 \pm 0.06$ \\
\hline

\end{tabular}
\end{center}
\label{tab:res}
\end{table}

There are several conclusions to be drawn from the results given in Table~\ref{tab:res}. First, as shown in Section~\ref{sec:genper}, generic policies make use of data that comes from different domains and this improves performance over an in-domain baseline, even in the case presented here where the domains are very different. The multi-policy MBCM results in performance which is either significantly better than other methods or statistically indistinguishable from other methods. In the case of limited training data, its performance is at least as good as the gold standard\footnote{Single generic policy (GEN) was not investigated on a larger training set since the main reason for having a generic policy would be to boost the performance when there is limited data available. In that case, the multi-policy Bayesian committee machine (MBCM) performs better.}. Another advantage of MBCM is that it does not require storing a separate generic policy model but only ever produces in-domain models that have the ability to contribute to action-selection for other domains.

Unlike domain-independent policy models~\cite{wwss15}, MBCM allows flexible selection of committee members. The usefulness of each committee member in the MBCM multi-policy model is explored in Table~\ref{tab:committee} for the SFR domain. As can be seen from the results, all committee members contribute to performance gains. However not all committee members are equally important. In this case, for good performance on the SFR domain, the SFH committee member is more useful than the L11 committee member.

\begin{table}[htbp]
\caption{Selection of committee members for multi-policy Bayesian committee machine for SFR domain. The committee policy is trained on 7500 dialogues equally spread across three domains.} 
\begin{center}
\begin{tabular}{ l | r | r | r  } 
\hline
\multicolumn{4}{c}{MBCM -- SFR  }  \\
\hline
Committee                     &         Reward  &   Success  &   \#Turns \\
 members                     &           &    &   \\
\hline
SFR   & $   7.32 \pm 0.22$ & $  79.97 \pm 0.82$ & $   8.51 \pm 0.10$ \\
SFR+SFH  & $   9.20 \pm 0.18$ & $  86.51 \pm 0.70$ & $   8.05 \pm 0.09$ \\
SFR+L11  & $   8.73 \pm 0.19$ & $  84.56 \pm 0.73$ & $   8.12 \pm 0.09$ \\
SFR+SFH+L11   & $   9.67 \pm 0.17$ & $  88.28 \pm 0.66$ & $   7.96 \pm 0.08$ \\
\hline

\end{tabular}

\label{tab:committee}
\end{center}
\end{table}

\subsection{Policy committee evaluation with human users}
\label{sec:policycommitteereal}

In order to fully examine the effectiveness of the proposed adaptation scheme, policies were also trained in direct interaction with human users. We compare two set-ups: one where an in-domain L6 policy is trained on-line and another where a multi-policy Bayesian committee machine is trained from scratch using data from the SFR, SFH and L6 domains, which produces a policy committee which can operate on all three domains. To the best of our knowledge, this is the first time a dialogue policy has been trained on multiple domains on-line in interaction with real users.

\begin{figure}[ht!]

\centering
\includegraphics[trim = 0mm 5mm 0mm 5mm, width=100mm]{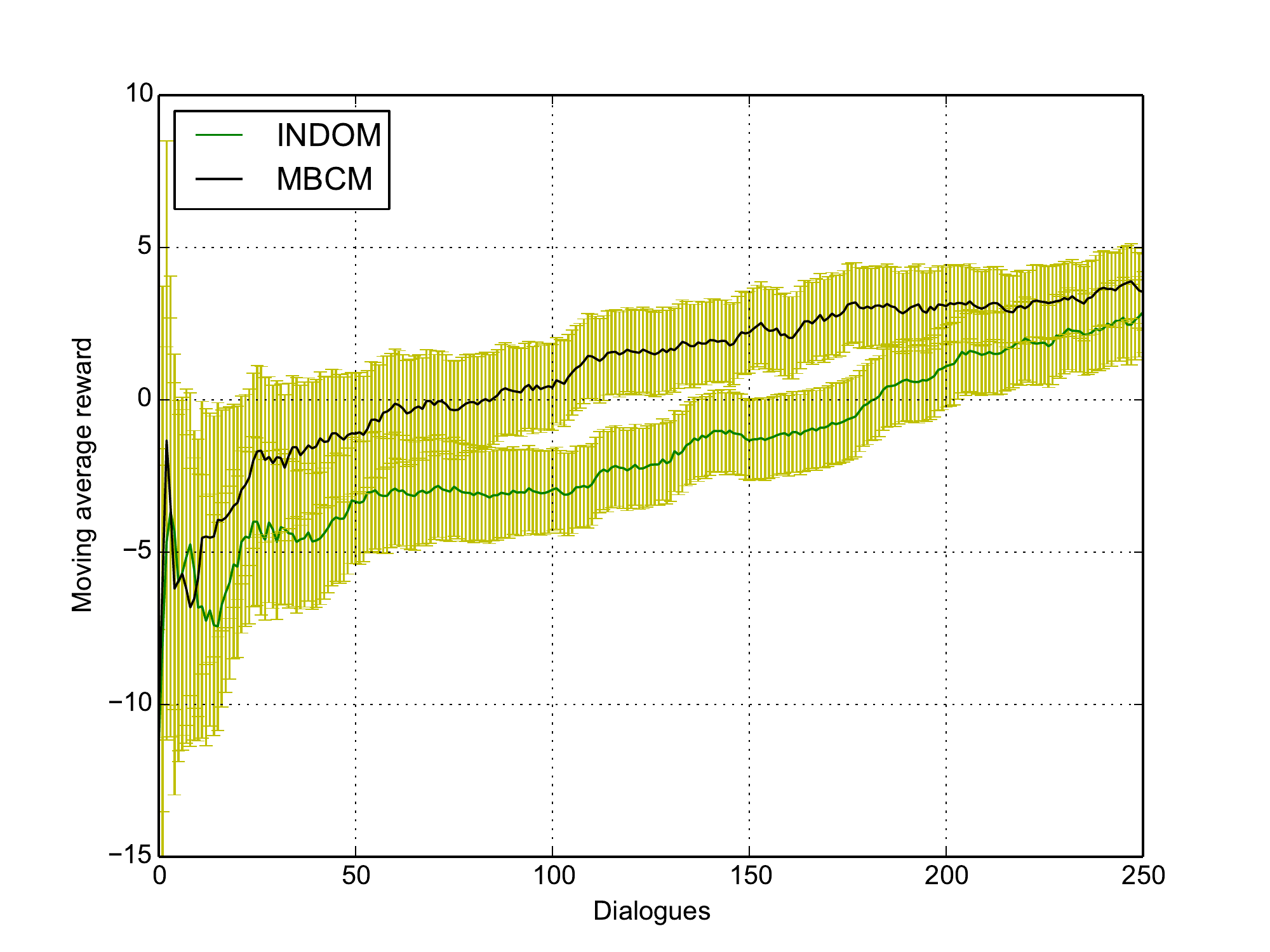}

\caption{Training in interaction with human users on L6 domain -- moving average reward. Plots are average of three sample runs.}
\label{fig:L6humtrarew}

\end{figure}

Fig.~\ref{fig:L6humtrarew} shows the moving average reward as a function of the number of training dialogues for the L6 domain comparing the in-domain (INDOM) policy and the multi-policy Bayesian committee machine (MBCM) as defined in Section~\ref{sec:mdds}. The performance of the MBCM policy was only shown on training dialogues that came from the L6 domain, but in fact it was also trained on SFR and SFH domains in parallel. The training data across the domains was equally distributed. Each plot is an average of three sample runs. The moving window was set to $100$ dialogues so that after the initial $100$ dialogues each point on the graph is an average of $300$ dialogues. The shaded area represents a 95\% confidence interval.   The initial parts of the graph exhibit more randomness in behaviour because the number of training dialogues is small. The results show that the multi-policy Bayesian committee machine consistently outperforms the in-domain policy. The caveat is that the computational complexity linearly increases with the number of committee members. Therefore this method would require a technique that selects and removes committee members as needed. An exploration of such a method goes beyond the scope of this article.

\subsection{Multi-agent simulation results}\label{sec:multiagentsim}

Finally, we examine the effectiveness of extending the policy committee model to multi-agent learning. The contrasts studied were as follows:

\begin{description}
\item{NA\"{I}V} \textbf{Na\"{i}ve approach} -- The total reward is given to each committee member during every interaction regardless of the current domain.
\item{WINN} \textbf{Winner-takes-all approach} -- The total reward is given to the policy member which on average gave the highest $Q$-value $Q$-variance ratio during the whole dialogue, $\Sigma^Q_{i}(\bm{b},a)^{-1}\overline{Q}_{i}(\bm{b},a)$ from Eq.~\ref{eq:bcm}, for each action taken by the system. 
%\item{SELFSCALE} \textbf{Scale received reward according to self $Q$-value estimate} -- Each policy committee member scales the total reward according to the average portion of its associated $Q$-value for the action that the system took as opposed to other actions it proposed.
\item{SCALE} \textbf{Scale received reward according to all committee members' $Q$-value estimate} -- the total reward is distributed to each policy committee member in proportion to the average $Q$-value $Q$-variance ratio, $\Sigma^Q_{i}(\bm{b},a)^{-1}\overline{Q}_{i}(\bm{b},a)$ from Eq.~\ref{eq:bcm}, for each action that the system took relative to the $Q$-value $Q$-variance ratios of the other committee members for the taken action.
\item{MBCM} \textbf{Multi-policy Bayesian committee machine} -- Each committee member is trained only on in-domain data, so the reward is passed only to the committee member which is specific to that domain (see Section~\ref{sec:mdds} for details).
\end{description}
 
We consider a multi-domain system for SFR, SFH and L11.  Two scenarios are examined: (a) when the training data is limited, with only $250$ dialogues available for each domain, and (b) when there is more training data available, $2500$ for each domain. 

For each method described above, 10 policies were trained on the simulated user using different random seeds. Each policy was then evaluated using $1000$ dialogues on each domain.   The overall average reward, success rate and  number of turns are given in Table~\ref{tab:res2} together with 95\% confidence intervals. We do not report results on SFH domain as policies on this domain behave similarly to the ones on SFR domain (see Fig.~\ref{fig:trainingSFRestaurant}~and~\ref{fig:trainingSFHotel}).

\begin{table}[htbp]
\centering
\caption{Comparison of strategies for multi-domain adaptation. In-domain performance is measured in terms of reward, success rate and the average number of turns per dialogue. Results are given with 95\% confidence intervals. } 
\begin{center}
\begin{tabular}{ l | r | r | r  } 
\hline
Strategy                     &         Reward  &   Success  &   \#Turns \\
\hline
\hline
\multicolumn{4}{c}{SFR trained on $750$ dialogues from SFR, SFH, L11 }  \\
\hline
NA\"{I}V  & $   7.00 \pm 0.20$ & $  73.66 \pm 0.86$ & $   7.70 \pm 0.08$ \\
WINN  & $   6.84 \pm 0.21$ & $  75.81 \pm 0.84$ & $   8.29 \pm 0.09$ \\
%SELFSCALE & $   6.86 \pm 0.20$ & $  72.90 \pm 0.87$ & $   7.68 \pm 0.08$ \\
SCALE  & $   7.06 \pm 0.21$ & $  75.29 \pm 0.85$ & $   7.98 \pm 0.09$ \\
MBCM  & $   7.37 \pm 0.20$ & $  76.60 \pm 0.83$ & $   7.92 \pm 0.08$ \\
\hline

\multicolumn{4}{c}{L11 trained on $750$ dialogues from SFR, SFH, L11 }  \\
\hline
NA\"{I}V  & $   8.82 \pm 0.20$ & $  77.40 \pm 0.82$ & $   6.63 \pm 0.07$ \\
WINN& $   7.23 \pm 0.22$ & $  72.35 \pm 0.88$ & $   7.20 \pm 0.09$ \\
%SELFSCALE   & $   7.69 \pm 0.21$ & $  72.17 \pm 0.88$ & $   6.72 \pm 0.07$ \\
SCALE  & $   8.11 \pm 0.21$ & $  74.61 \pm 0.85$ & $   6.78 \pm 0.08$ \\
MBCM  & $   8.52 \pm 0.20$ & $  77.09 \pm 0.82$ & $   6.88 \pm 0.07$ \\
\hline
\hline
\multicolumn{4}{c}{SFR trained on $7500$ dialogues from SFR, SFH, L11 }  \\
\hline
NA\"{I}V  & $   9.45 \pm 0.22$ & $  87.98 \pm 0.85$ & $   8.14 \pm 0.11$ \\
WINN & $   9.67 \pm 0.18$ & $  89.24 \pm 0.68$ & $   8.15 \pm 0.09$ \\
SCALE  & $   9.41 \pm 0.17$ & $  88.08 \pm 0.66$ & $   8.18 \pm 0.09$ \\
MBCM & $   9.67 \pm 0.17$ & $  88.28 \pm 0.66$ & $   7.96 \pm 0.08$ \\

\hline

\multicolumn{4}{c}{L11 trained on $7500$ dialogues from SFR, SFH, L11 }  \\
\hline
NA\"{I}V&  $  10.92 \pm 0.16$ & $  86.80 \pm 0.70$ & $   6.42 \pm 0.07$ \\
WINN & $  11.25 \pm 0.18$ & $  88.51 \pm 0.76$ & $   6.43 \pm 0.08$ \\
SCALE & $  11.24 \pm 0.17$ & $  88.55 \pm 0.69$ & $   6.44 \pm 0.07$ \\
MBCM& $  10.73 \pm 0.16$ & $  87.23 \pm 0.66$ & $   6.70 \pm 0.07$ \\
\hline
\hline
\multicolumn{4}{c}{Averaged across domains and size of training data}  \\
\hline
NA\"{I}V & $   8.94 \pm 0.10$ & $  80.49 \pm 0.42$ & $   7.13 \pm 0.04$ \\
WINN & $   8.46 \pm 0.10$ & $  80.37 \pm 0.42$ & $   7.58 \pm 0.05$ \\
SCALE & $   8.83 \pm 0.10$ & $  81.17 \pm 0.40$ & $   7.38 \pm 0.04$ \\
MBCM & $   9.06 \pm 0.09$ & $  82.17 \pm 0.38$ & $   7.35 \pm 0.04$ \\
\hline

\end{tabular}

\label{tab:res2}
\end{center}
\end{table}

Conclusions that can be drawn from these results are the following. First, on a smaller dataset the WINN approach, which chooses the winning committee member to pass the total reward to, is less effective than the approaches which distribute the reward. This is expected, as in the latter case the agent's policy learns from a larger set of dialogues, which is particularly useful in the early stages of the optimisation process. On larger datasets, the winner-takes-all approach gives similar or better performance to the approaches which distribute the reward. This means that if large amount of data is available one can afford to use the model which chooses a which subset of data to optimise the policy. In this case that is the data which has the most accurate reward estimate. If we average results across the domains and the sizes of the training data, however, we can see that it is generally more effective to use the approaches which distribute the reward. 

It is also important to understand behaviour when a new domain is added alongside a set of existing agents which themselves are not yet fully trained.  We are interested in both the performance in the new domain as well as the existing domains.
To investigate this, two agents operating in the SFR and SFH domains were pre-trained with 250 dialogues each using SCALE reward distribution mechanism. Performance was then evaluated in the SFR and the new as yet untrained L11 domain. The L11 agent was then trained with 250 dialogues in that domain. Again, the performance was tested in both L11 and SFR. Finally training continued with another 250 dialogues for each of the three domains - SFR, SFH and L11 and the performance in the SFR and L11 domains tested for a final time. The results are shown in Table~\ref{tab:dynagent}. 

\begin{table}[htbp]
\caption{Performance when adding a new agent for the L11 domain to a multi-domain dialogue manager using SCALE training with two partially trained agents for the SFR and SFH domains.}
\begin{center}
\begin{tabular}{ c|r|r|r } 
\hline
\multicolumn{4}{c}{Performance in L11 domain}  \\
\hline
Training data & Reward & Success & Turns\\
\hline
250 SFR+ 250 SFH& $ -10.89 \pm 0.40$ & $  39.89 \pm 0.96$ & $  16.65 \pm 0.21$\\
+250 L11& $   4.18 \pm 0.28$ & $  62.18 \pm 0.95$ & $   7.89 \pm 0.11$ \\ 
+250 SFR+250 SFH+250 L11 & $   7.26 \pm 0.22$ & $  70.47 \pm 0.94$ & $   6.79 \pm 0.08$  \\ 
\hline
\multicolumn{4}{c}{Performance in SFR domain}  \\
\hline
250 SFR+ 250 SFH& $   6.12 \pm 0.22$ & $  70.22 \pm 0.91$ & $   7.90 \pm 0.08$ \\
+250 L11& $   6.75 \pm 0.21$ & $  73.54 \pm 0.86$ & $   7.93 \pm 0.08$ \\ 
+250 SFR+250 SFH+250 L11 & $   8.05 \pm 0.20$ & $  79.38 \pm 0.83$ & $   7.79 \pm 0.08$ \\
\hline
\end{tabular}
\end{center}\label{tab:dynagent}
\end{table}

The performance of the dialogue manager in the L11 domain when trained only with SFR and SFH dialogues is very poor, which is expected as these are very different domains. However with the addition of 250 L11 dialogues, the performance dramatically improves. What is more, adding these L11 dialogues does not impede performance in the SFR domain, in fact it improves slightly. With an additional 750 dialogues spread across all three domains, the performance significantly improves in both the L11 and SFR domains.

\subsection{Multi-agent human user evaluation}
\label{sec:multiagentreal}

To ensure that the benefits of the proposed reward distribution approach suggested by the above simulation results carry over into systems trained
on-line, two systems were also trained in direct interaction with human users. First, a multi-policy Bayesian committee machine~(MBCM) was trained from scratch using data from the SFR restaurant, the SFH hotel and the L6 laptop domains. This MBCM policy committee machine operates on all three domains but is dependent on the knowledge of the current domain for policy updating. This is compared to the committee reward scaling~(SCALE) machine, presented in Section~\ref{sec:multiagentsim}, which distributes the reward to every committee member for each dialogue regardless of the domain. The system was deployed in a telephone-based set-up, with subjects recruited via Amazon MTurk and a recurrent neural network model was used to estimate the reward ~\cite{svgk15}.

\begin{figure}[ht!]

\centering
\includegraphics[trim = 0mm 5mm 0mm 5mm, width=100mm]{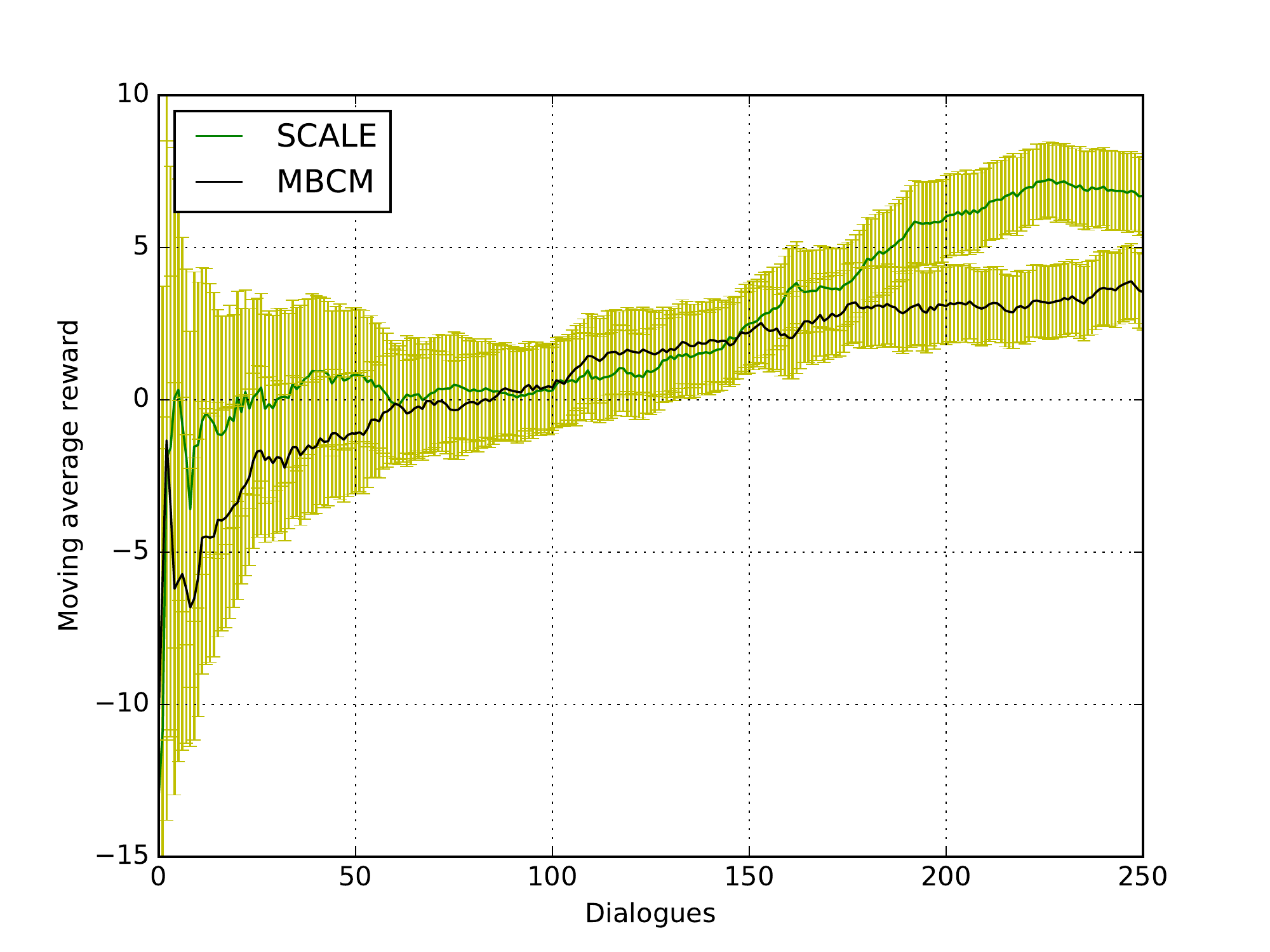}

\caption{Training using the MBCM and SCALE approaches in interaction with human users in the L6 laptop domain -- moving average reward. Plots are average of three sample runs.}
\label{fig:mbcmhumtrarew}

\end{figure}

Fig.~\ref{fig:mbcmhumtrarew} shows the moving average reward as a function of the number of training dialogues for the L6 domain comparing the MBCM and SCALE committee approaches. The committees were also trained on SFR and SFH domains in parallel. The training data across the domains was equally distributed. As in Section~\ref{sec:policycommitteereal} each plot is an average of three sample runs. The moving window was set to $100$ dialogues so that after the initial $100$ dialogues each point on the graph is an average of $300$ dialogues. The shaded area represents a 95\% confidence interval. As can be seen from the reward graph for the SCALE approach, the results confirm that it is not necessary for the committee to be aware of the domain.  On the contrary, distributing reward to each committee member according to their contribution can even produce better performance than only sending the reward signal to the committee member dedicated to the current domain.

\section{Conclusion}\label{sec:concl}

This paper has described three models which support dialogue system domain extension. First,  a  distributed multi-domain dialogue architecture was proposed in which dialogue policies are organised in a class hierarchy aligned to an underlying knowledge graph.  The class hierarchy allows a system to be deployed by using a modest amount of data to train a small set of generic policies.  As further data is collected, generic policies can be adapted to give in-domain performance.  Using Gaussian process-based reinforcement learning, it has been shown that it is possible to construct generic policies which provide acceptable in-domain user performance, and better performance than can be obtained using under-trained domain specific policies.   To construct a generic policy,  a design consisting of all common slots plus a number of abstract slots which can be mapped to domain-specific slots works well. It has also been shown that as sufficient in-domain data becomes available, it is possible to seamlessly adapt to improve performance, without subjecting users to unacceptable disruptions in performance during the adaptation period and without limiting the final performance compared to policies trained from scratch. 

An alternative to hierarchically structured policies is the distributed committee model which uses estimates from different policies for action selection at every dialogue turn. The results presented have shown that this model is particularly useful for  training multi-domain dialogue systems where the data is limited and varied. As shown in both simulations and in real user trials, the Bayesian policy committee approach gives superior performance to the traditional one-policy-approach across multiple domains and allows flexible selection of committee members during testing.

Finally, the basic policy committee model was extended using  ideas from multi-agent learning to distribute the reward signal among the committee members.  This model is particularly useful in real-world scenarios where the domain is a priori unknown and indeed, may change during a dialogue. In simulations, the proposed approach achieves a performance which is close to that which relies on explicit domain information to assign reward, while in a real human trial, it produced better performance.

For future work, these methods will be applied to a dialogue manager operating over a large knowledge graph in order to demonstrate that they do
indeed scale and offer a viable approach to building truly open domain spoken dialogue systems which learn on-line in interaction with real users.

\section{Acknowledgements}

The research leading to this work was funded by the EPSRC grant EP/M018946/1 "Open Domain Statistical Spoken Dialogue Systems". The data used in these experiments is available at https://www.repository.cam.ac.uk/handle/1810/259963.




\bibliographystyle{elsarticle-num}
\bibliography{refs}
\end{document}